# GuLu·XuanYuan , a biomimetic Transformer that intergrates humanoid MIP, reptile UGV, and bird UAV

Le Chen*[1] | Jie Yu[2] | XingWu Chen[3]

[1] School of Computer and Mathematics, Fujian University of Technology, Fujian, China

[2] School of Mechanical and Automotive Engineering, Fujian University of Technology, Fujian, China

[3] School of Electronics and Physics, Fujian University of Technology, Fujian, China

**Correspondence**
*Le Chen, Email: lechen@fjut.edu.cn

**Summary**

This article proposes a multi habitat bio-mimetic robot, named as GuLu·XuanYuan. It combines all common types of mobile robots, namely humanoid MIP, unmanned ground vehicle, and unmanned aerial vehicle. These 3 modals imitate human, bird, and reptile, separately. As a transformer, GuLu·XuanYuan can transform from one modal to another. Transforming function integrates the specialized abilities of three robots into the same machine body. This simplification approach helps to reduce the total number of required robots. From another perspective, the deformation function is equivalent to creating more economic value.

## 1 | INTRODUCTION

In 1984, Takara Tomy and Hasbro jointly released a science fiction cartoon, "Transformers". Afterwards, they have been commit- ted to developing plastic toys based on robot characters in their cartoon. However, 40 years have passed and the real multi-morpho robot in science fiction cartoon still cannot be achieved. A few months ago, Takara Tomy developed a lunar transformer named SORA-Q, which looked like a baseball with a diameter of 8 centimeters rolling on the surface of the moon. Developing larger transformer is very hard. We estimate that the performance limit of current motors can only drive deformable robots within a height range of 0.5 to 2 meters. Therefore, humanoid robots with such height are more suitable for transforming into unmanned ground vehicles (UGV) and unmanned aerial vehicles (UAV), rather than cars or planes that can carry people like the Autobots and Decepticons in science fiction cartoon.

At present, the best work on transforming between UGV and UAV is Multi-Modal Mobility Morphobot (M4) developed by Caltech. We improved M4 and proposed a new transformer and named it GuLu·XuanYuan. Owing to the longer limbs and a flexible chassis with spinal structure, GuLu·XuanYuan pioneered true transforming between humanoid and UAV. We wish it was able to integrate the abilities of various specialized robots such as UGV, UAV, and humanoid MIP and meet the comprehensive requirements of diverse and complex indoor/outdoor rescue, accompanying the elderly and young and even complete exoplanet missions, such as barren terrain exploration, lightweight item handling, and daily maintenance of extraterrestrial bases. After all, at current stage, the launch cost of interstellar carrier rocket remains very high. It is rumored that NASA's shipping cost for supplying materials to the space station is as high as $10000 to $100000 per kilogram. If a deformable robot can undertake the tasks of three types of mobile robots in outer space or even on outer planets, then the aerospace department can significantly reduce the cost of interplanetary transportation for robots.

## 2 | RELATED WORKS

Within the scope of our knowledge, only ETH Zurich claims to have initially achieved the deformation ability of humanoid robots, while Caltech, Ben-Gurion U., and Harbin I.T. are more inclined to study the problem that how to transform UAV into UGV.1 2ANYmal, the famous wheeled robotic dog by ETH, can stand upright and then clamp a box with a pair of front wheeled limb.3 2This idea was also mentioned in the General Theory of Multi-Modal Mobility Morphobot (M4) at the California Institute of Technology in the United States.4M4 has its innovative wheeled legs, which is an important improvement on the wheeled legs of the crawling drone "Flying Star".1 Sihite and Kalantari found that both UGV wheel and UAV propeller occupy samilar cake space and are therefore very suitable for integration by stacking on the central axis of a same circular metal frame. This solution visibly reduces the complexity of the overall mechanical structure of the robot and also greatly saves the occupation space.

## 3 | SOLUTION

After decades of development, small robots have finally condensed into three common forms: humanoid,UAV, and UGV. Sihite et al. first proposed the idea of unifying the aforementioned types of mobile robots. They proposed M4 as an example. They adopted standard and simple structures of UGV and UAV, which resulted in its humanoid shape very poor and far fetched. Therefore, we speculated that there should be some additional components in the machine structures of UAV and UGV, which could serve as missing limbs for the M4 humanoid robot. In other words, the design of GuLuXuanYuan should refer to some more complex UAV & UGV. After analyzing some recently emerged robots, we identified three potentially suitable biomimetic robots: robot dog-human ANYMal that ofen walks upright, robot lizard that crawls on four legs, and robot bird SNAG that flies over the forest and perches on high branches. We compared the mechanical structures of the three classic biomimetic machines and identified similarities in their components. Then, we redesigned these similar components so that they can be used in all three forms. The remaining parts were designed as intermediaries connecting the above crucial components and their positions in each robot structure were difficult to change. Finally, we derived a deformation solution for the GuLu·XuanYuan based on the positions of these core components, as shown in Figure 1.

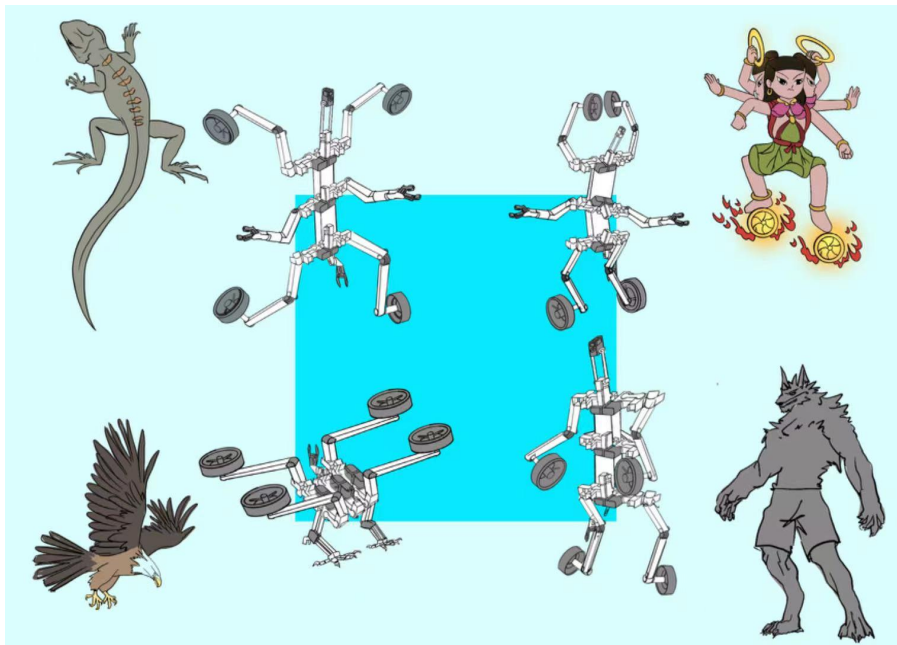

**FIGURE 1** The deformation solution .

GuLu·XuanYuan certainly has all the functions of M4, but its working behavior is quite different. M4 is a biomimetic com- bination of bird (wings and legs), mongooses, and seals, while GuLu·XuanYuan integrates hawk (claws), quadruped-humanoid (dog) and lizard. The new accessory components we introduced mainly include 3 parts, a pair of large robotic arms, 4 knees (or 2 knees and 2 elbows), and a spinal structure in the UGV chassis. In addition, we increased the length of the wheel legs as well. These improvements enable GuLu·XuanYuan to undertake more specific tasks, rather than only 3 most basic modal func- tions of M4, running of UGV, flying of UAV, standing of MIP. Let's analyze in detail the roles played by these newly introduced components in the three modes.

- When GuLu·XuanYuan stands up from the crawling state of UGV, it has four robotic arms available for its humanoid mode to choose from. If we command the quadruped animal's forelimbs serve as humanoid arms, the working principle of GuLu·XuanYuan is very similar to the quadruped-humanoid mode of robotic dog ANYmal, shown in Figure 2 (right). Since it is a standing upright beast, we called this humanoid modal as a werewolf, shown in Figure 1 (bottom-right). If GuLu·XuanYuan uses a pair of large robotic arms to work and extends a pair of small robotic wheel arms towards the back, then we call this humanoid form Nezha, who was a young superhero in Chinese mythology, shown in Figure 2 (left). Because they look very similar. Nezha had three heads and six arms. He stepped on fire wheels with its feet and holds two big metallic rings as weapons in his hands, shown in Figure 1 (upper right). Both werewolf and Nezha have a pair of idle robotic arms. In fact, the idle arms are not redundant items. We throw the idle machine arms to the robot's back to serve as the backpack. Maybe it is Boston Dynamics that first adopted such backpack technology, which can balance the robot Handle's body to prevent it from falling when its center of gravity is about to shift.

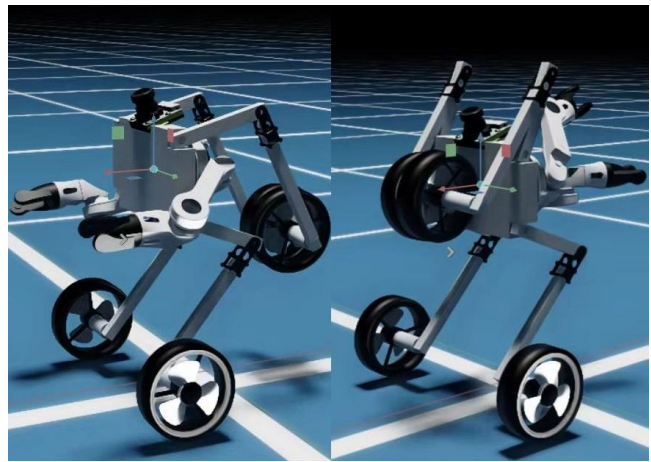

**FIGURE 2** Humanoid MIP.

- The crucial components of the bird modal play the roles of four propellers and a pair of mechanical bird claw, as shown in Figure 3. Its propeller is a combination of M4 propeller-wheel and SNAG propeller, as shown Figure 3.5 The pure mechanical claws of SNAG grasp tree branches passively, while the claws of GuLu·XuanYuan operates actively and can do other jobs.

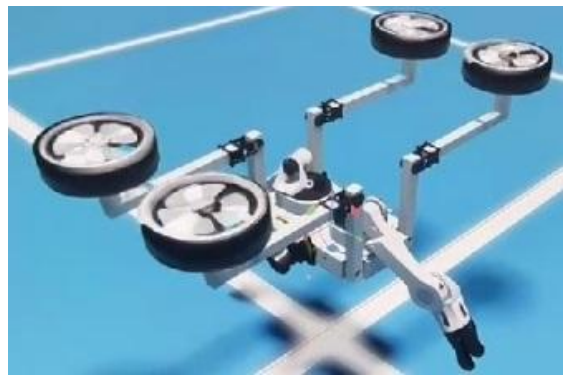

**FIGURE 3** Bird UAV.

- Compared to the autonomous vehicle form of the M4, GuLu·XuanYuan has longer wheeled legs. Thus, its UGV modal is much closer to a Mars rover, as shown in Figure 4. Its lookis like the famous robot dog ANYmal, except for the two big robotic legs. In our opinion, it is rather a robotic reptile than either a dog or a spider.[6,7] Because both ANYmal and robotic spiders have no spinal structure, while Xuanyuan does have. With the coordination of spinal joints, the Mars rover modal of GuLu·XuanYuan can crawl almost prostrate on the rugged surface of alien planets. When the body was going to roll over, the mechanical reptile's legs can provide support and enhance the driving force of the Mars rover's crawling.

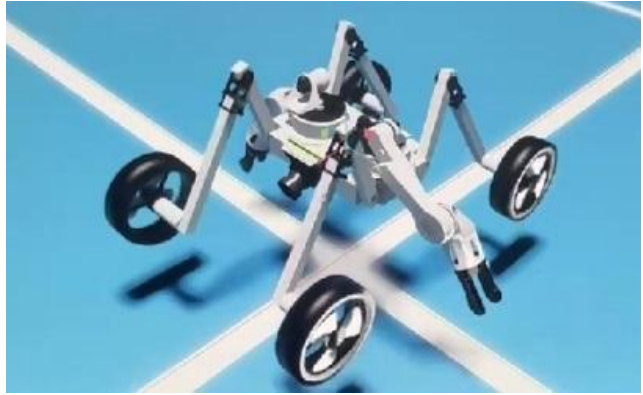

**FIGURE 4** Reptile UGV.

## 4 | EXPERIMENT

After conducting some simulations using Nvidia Isaac Sim, we began experiments to fabricate and test the real robot. We will show the progress of the experiment in the future.

## ACKNOWLEDGMENTS

The first author thanks graduate students Xiang Zhang and Yihui Wang for their efforts in building the Xuan Yuan model using Solidworks. The first author also thanks Linjing Zhang for drawing Figure 1.

Le Chen ET AL | 57. Bing Z, Rohregger A, Walter F, Huang Y, Lucas P, Morin FO, et al. Lateral flexion of a compliant spine improves motor performance in a bioinspired mouse robot. Science Robotics. 2023;**8**(85):eadg7165.